\documentclass[conference]{IEEEtran}
\IEEEoverridecommandlockouts
\usepackage{cite}
\usepackage{amsmath,amssymb,amsfonts}
\usepackage{algorithmic}
\usepackage{graphicx}
\usepackage{textcomp}
\usepackage{xcolor}
\usepackage[a4paper, total={184mm,239mm}]{geometry}
\def\BibTeX{{\rm B\kern-.05em{\sc i\kern-.025em b}\kern-.08em
    T\kern-.1667em\lower.7ex\hbox{E}\kern-.125emX}}
    
\usepackage{amsthm}
\usepackage{algorithm}
\usepackage{multirow}

\newtheorem{problem}{Problem}

\newtheorem{definition}{Definition}    

\begin{document}

\title{Efficient Global Robustness Certification of Neural Networks via Interleaving Twin-Network Encoding
\thanks{This work is supported in part by NSF grants 1834701, 1839511, 1724341, 2038853, and ONR grant N00014-19-1-2496.}
\thanks{This paper is published on \emph{Design, Automation and Test in Europe Conference (DATE 2022)}.}
} 
\author{
\IEEEauthorblockN{Zhilu Wang\IEEEauthorrefmark{1},
Chao Huang\IEEEauthorrefmark{2},
Qi Zhu\IEEEauthorrefmark{1}}
\IEEEauthorblockA{\IEEEauthorrefmark{1}Department of Electrical and Computer Engineering, Northwestern University, Evanston, IL, USA\\
Emails: zhilu.wang@u.northwestern.edu, qzhu@northwestern.edu}
\IEEEauthorblockA{\IEEEauthorrefmark{2}Department of Computer Science, University of Liverpool, Liverpool, UK\\
Email: chao.huang2@liverpool.ac.uk
}}

\maketitle

\begin{abstract}
The robustness of deep neural networks has received significant interest recently, especially when being deployed in safety-critical systems, as it is important to analyze how sensitive the model output is under input perturbations. While most previous works focused on the \emph{local robustness} property around an input sample, the studies of the \emph{global robustness} property, which bounds the maximum output change under perturbations over the entire input space, are still lacking.
In this work, we formulate the global robustness certification for neural networks with ReLU activation functions as a mixed-integer linear programming (MILP) problem, and present an efficient approach to address it. Our approach includes a novel interleaving twin-network encoding scheme, where two copies of the neural network are encoded side-by-side with extra interleaving dependencies added between them, and an over-approximation algorithm leveraging relaxation and refinement techniques to reduce complexity.  
Experiments demonstrate the timing efficiency of our work when compared with previous global robustness certification methods and the tightness of our over-approximation.
A case study of closed-loop control safety verification is conducted, and demonstrates the importance and practicality of our approach for certifying the global robustness of neural networks in safety-critical systems.
\end{abstract}

\section{Introduction}
Deep neural networks (DNNs) are being applied to a wide range of tasks, such as perception, prediction, planning, control, and general decision making. While DNNs often show
significant advantages in performance over traditional model-based methods, pressing concerns have been raised on the uncertain behaviors of DNNs under varying inputs, especially for safety-critical systems such as autonomous vehicles and robots. 
Some of the well-trained DNNs are found to be vulnerable to a small adversarial perturbation on their inputs~\cite{Biggio_2013_ECML}.
The \emph{robustness} metric of DNNs is defined to bound such uncertain behaviors when the input is perturbed. Specifically, the robustness of a DNN represents how much its output varies when its input has a bounded perturbation, where the perturbation can be either from random noises or due to malicious attacks.

Formal methods, such as Satisfiability Modulo Theories (SMT)~\cite{Katz_2017_CAV_Reluplex,Huang_2017_CAV} and Mixed-Integer Linear Programming (MILP)~\cite{Lomuscio_2017-arxiv}, have been used to either find an adversarial example or certify the robustness (i.e., proving no adversarial example exists) around a given input sample, which is considered as \emph{local robustness}.
The efficiency is improved by certifying a conservative (sound but not complete) local robustness bound via over-approximated output range analysis methods~\cite{Singh_2019_POPL,Huang_2020_EMSOFT}.

In safety-critical systems with DNNs involved, it is important to analyze the DNN robustness for evaluating system safety. As local robustness is defined locally to a given input sample, to evaluate system safety, we will need to apply local robustness certification \emph{during runtime} for each input sample that the system encounters or may encounter. However, the complexity of local robustness certification, even with conservative over-approximation, is too high for runtime execution in most practical systems. Moreover, sometimes it is required to ensure system safety for future time horizon (e.g., to verify there is no collision for the next 3 seconds). As future input samples are unknown, local robustness certification cannot be applied.

Instead, we could address system safety by considering the problem of \emph{global robustness}, which measures the worst-case DNN output change against input perturbation for all possible input samples. The definition of the global robustness is introduced in~\cite{Katz_2017_CAV_Reluplex}, with a proposed SMT-based tool Reluplex for certifying the exact global robustness. In~\cite{Cheng_2017_ATVA}, the global robustness for classification problems is formulated into MILP to find the maximum input perturbation that can preserve the classification result. In~\cite{Chen_2021_arxiv}, an MILP-based exact global robustness certification method is proposed for logic ensemble classifier (a generalized-version of decision tree, not DNN). 

A major challenge for exact global robustness certification is its complexity -- while it can be run offline to facilitate safety analysis (since it considers the entire input space), the high complexity often makes it intractable even in offline computation. A number of methods have been proposed to tackle the complexity, however they often lack the deterministic guarantees needed for safety verification.  
For instance, dataset-based global robustness estimation approaches are proposed in~\cite{Ruan_2019_IJCAI,Bastani_2016_NIPS}, which conduct local robustness certification for each sample in the dataset and find the worst case or the average. As the dataset cannot cover all possible input samples, the derived global robustness is just an estimation with no deterministic guarantees. 
Sampling-based methods~\cite{Mangal_2019_ICSE} randomly sample the DNN inputs, evaluate the local robustness of the random-sampled inputs, and provide a probabilistic global robustness. Similarly, they cannot provide deterministic guarantees. 
Region-based robustness analyses~\cite{Gopinath_2018_ATVA,Mangal_2019_ICSE} divide the input space into regions, where all possible inputs in each certified-robust region have the same classification result. However, the number of regions could be extremely large for DNNs with high-dimensional inputs, making it hard to certify all regions and provide deterministic guarantees.

In this work, we propose an efficient certification approach to \emph{over-approximate} the global robustness. The derived global robustness is sound and deterministic, the over-approximation is tight, and the approach can be used effectively for the purpose of safety verification. To achieve this, our approach introduces a novel network encoding structure, namely interleaving twin-network encoding, to compare two copies of the neural network side-by-side under different inputs, with extra interleaving dependencies added between them to improve efficiency. Our approach also includes over-approximation techniques based on network decomposition and LP (linear programming) relaxation, to further reduce the computation complexity.
To the best of our knowledge, our approach is the \emph{first global robustness over-approximation method that certifies the robustness among the entire input domain with sound and deterministic guarantee}.
Experiments show that our approach is much more efficient and scalable than the exact global robustness methods such as Reluplex~\cite{Katz_2017_CAV_Reluplex}, with tight over-approximation -- our approach can certify DNNs with more than 5k neurons in 5 hours while the exact methods cannot certify DNNs with more than 64 neurons in 1 day. To demonstrate the importance of global robustness and the practical application of our certification approach, a case study is conducted to verify the safety of a closed-loop control system with a vision-based perception DNN in the loop.

\section{Efficient Global Robustness Certification}

\subsection{Problem Formulation}

\begin{figure}[t]
\centering
\includegraphics[width=0.8\columnwidth]{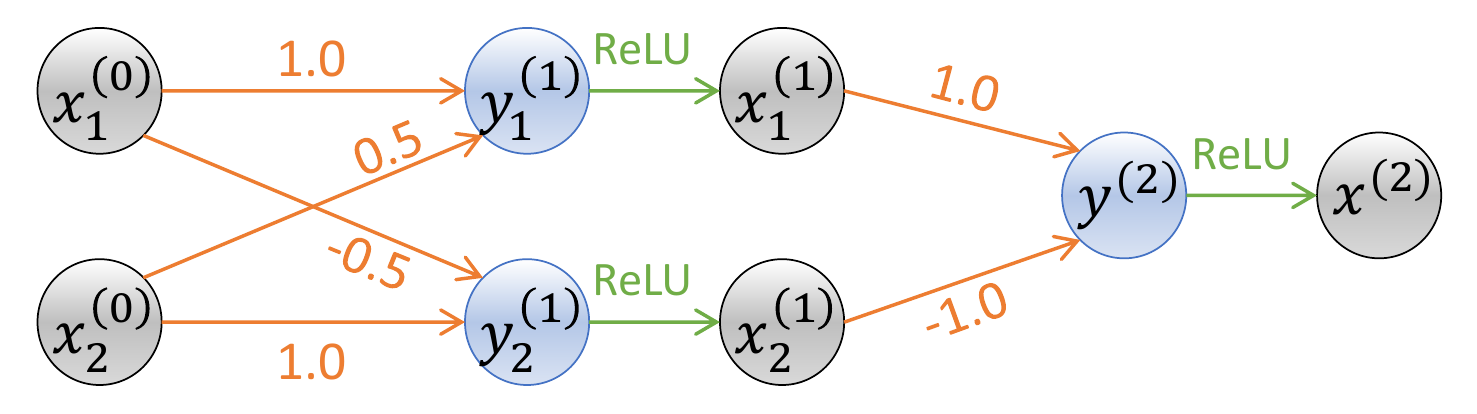}
\caption{An example neural network for illustration.}
\label{fig:NNstruct}
\vspace{-12pt}
\end{figure}

An $n$-layer neural network, formulated as $F:\mathbb{R}^{m_0}\rightarrow \mathbb{R}^{m_n}$ maps an $m_0$-dimension input $x^{(0)}\in\mathbb{R}^{m_0}$ into an $m_n$-dimension output $x^{(n)}\in \mathbb{R}^{m_n}$. 
In the neural network, the output of each layer $i$ is denoted as $x^{(i)}\in\mathbb{R}^{m_{i}}$ with dimension $m_i$, and layer $i$ maps the output from the previous layer $x^{(i-1)}$ into its output by $x^{(i)} = f^{(i)}(x^{(i-1)})$.
$f^{(i)}$ is composed with a linear transformation $y^{(i)}=W^{(i)} x^{(i-1)} + b^{(i)}$ (e.g., fully-connected layer, convolution layer, average pooling or normalization), and (optionally) a ReLU activation function. The linear transformation result is denoted as intermediate variable $y^{(i)}\in\mathbb{R}^{m_{i}}$. The layer output $x^{(i)}=relu(y^{(i)})=\max(y^{(i)}, 0)$ if it has ReLU activation function, and $x^{(i)}=y^{(i)}$ otherwise. An illustrating example is shown in Fig.~\ref{fig:NNstruct}, which is a 2-layer neural network that maps 2-dimension input $x^{(0)}$ into 1-dimension output $x^{(2)}$ with a 2-neuron hidden layer. For simplicity, the bias $b^{(i)}$ of each layer $i$ is set to 0.

We consider the global robustness over the entire input domain $X$ of the neural network, i.e., $\forall x^{(0)}\in X$. It measures the worst-case output variation when there is a small input perturbation for any possible input sample in the input domain $X$. Moreover, we focus on the input perturbation that is bounded in the form of $L_{\infty}$ norm, and have the same global robustness definition as~\cite{Katz_2017_CAV_Reluplex,Wang_2021_DATE}.

\begin{definition}[Global Robustness] \label{def:globRobust}
The $j$-th output of neural network $F$ is $(\delta, \varepsilon)$-globally robust in the input domain $X$ iff
\[
\forall x^{(0)}, \hat{x}^{(0)} \in X, \|\hat{x}^{(0)}-x^{(0)}\|_{\infty} \leq \delta \implies |\hat{x}^{(n)}_j - x^{(n)}_j|\leq \varepsilon
\]
where $x^{(n)}=F(x^{(0)})$ and $\hat{x}^{(n)} = F(\hat{x}^{(0)})$.
\end{definition}

In this work, we tackle the problem of \emph{how robust the neural network is, i.e., how small $\varepsilon$ can be for a given $\delta$,} formally as:
\begin{problem}\label{prob:globrobust}
For a neural network $F$, given an input perturbation bound $\delta$, determine the minimal output variation bound $\varepsilon$ such that $F$ is guaranteed to be $(\delta, \varepsilon)$-globally robust. 
\end{problem}

The work in~\cite{Katz_2017_CAV_Reluplex} proposes to solve the global robustness problem by encoding two copies of the neural network side by side, as illustrated in the left part of Fig.~\ref{fig:NNsidebyside}. The two network copies take two separate inputs $x^{(0)}$ and $\hat{x}^{(0)}$ and produce outputs $x^{(n)}$ and $\hat{x}^{(n)}$. As a perturbation of $x^{(0)}$, $\hat{x}^{(0)}$ is restricted by the input perturbation $\Delta x^{(0)} = \hat{x}^{(0)} - x^{(0)}$, where $\|\Delta x^{(0)}\|_\infty \leq \delta$. The output variation bound $\varepsilon$ is the bound of the output distance $\Delta x^{(n)} = \hat{x}^{(n)} - x^{(n)}$. Under this encoding, Problem~\ref{prob:globrobust} can be formulated as an optimization problem:
\begin{equation}
    \begin{aligned}
    \varepsilon := \max \quad & |\hat{x}^{(n)} - x^{(n)}|\\
    \textrm{s.t.} \quad 
    & \hat{x}^{(n)} = F(\hat{x}^{(0)}),\ x^{(n)}=F(x^{(0)})\\
    & \hat{x}^{(0)}, x^{(0)}\in X,\ \|\hat{x}^{(0)} - x^{(0)}\|_\infty < \delta
    \end{aligned}\label{equ:probformul}
\end{equation}

Eq.~\eqref{equ:probformul} can be solved by mixed-integer linear programming (MILP),
where the neural network $F$ can be decomposed into a series of neuron-wise dependency. For the $j$-th neuron of layer $i$ with ReLU activation, we have:
\begin{gather}
    y^{(i)}_j = W^{(i)}_j x_{i-1} + b^{(i)}_j,\quad
    x^{(i)}_j = \max (y^{(i)}_j, 0) \label{equ:milp_x}\\
    \hat{y}^{(i)}_j = W^{(i)}_j\hat{x}^{(i-1)} + b^{(i)}_j,\quad
    \hat{x}^{(i)}_j = \max (\hat{y}^{(i)}_j, 0) \label{equ:milp_x_2}
\end{gather}
The $\max(x, 0)$ function for each ReLU activation can be linearized by Big-M method~\cite{Cheng_2017_ATVA} with the introduction of a new integer (binary) variable to indication whether $y<0$.

The complexity to solve this MILP is exponential to the number of ReLU neurons, making it too complex for most practical neural networks. 
In this work, to overcome this challenge, we present 1) a new interleaving twin-network encoding (ITNE) scheme that adds extra interleaving dependencies between the two copies of the neural network, and 2) two approximation techniques leveraging the new encoding scheme, namely \emph{network decomposition (ND)} and \emph{LP relaxation (LPR)}, to efficiently find an over-approximated sub-optimal solution $\Bar{\varepsilon}$ of the output variation bound $\varepsilon$, where $\Bar{\varepsilon} \geq \varepsilon$.

\subsection{Interleaving Twin-Network Encoding} \label{sec:sbs_copies}
\begin{figure}[t]
\centering
\includegraphics[width=0.8\columnwidth]{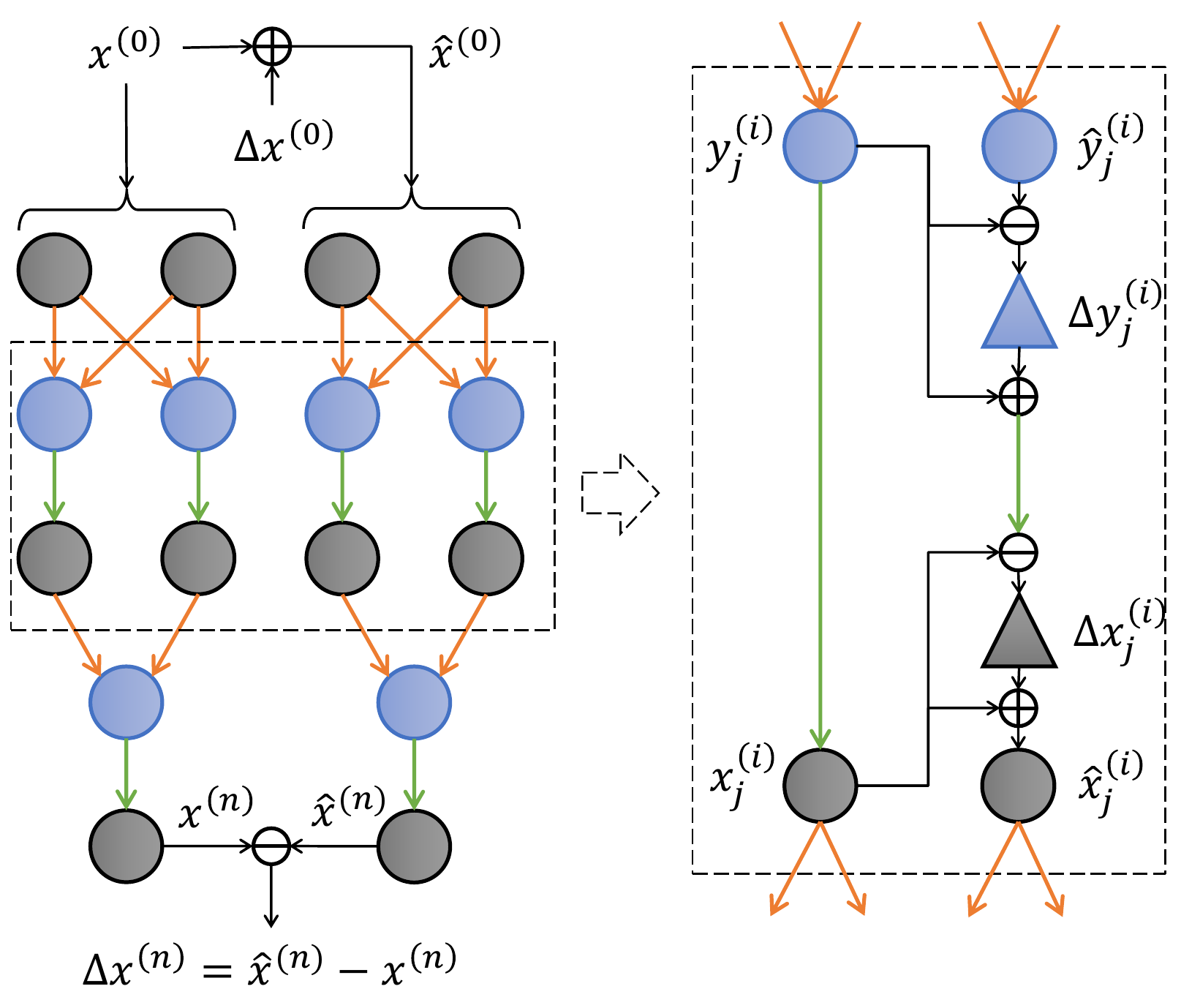}
\caption{Left: The basic twin-network encoding (BTNE) for global robustness certification. Right: the neuron-level interleaving twin-network encoding (ITNE) built upon the basic structure, where the hidden layer neurons are connected between the two copies with distance variables $\Delta y^{(i)}_j$ and $\Delta x^{(i)}_j$.}
\label{fig:NNsidebyside}
\vspace{-12pt}
\end{figure}

We call the neural network encoding scheme introduced in~\cite{Katz_2017_CAV_Reluplex}, shown in the left side of Fig.~\ref{fig:NNsidebyside} with only input and output layers connected, the basic twin-network encoding (BTNE). In contrast, the new network encoding scheme we designed to enable the over-approximation techniques for global robustness certification is shown in the right side of Fig.~\ref{fig:NNsidebyside} and called the interleaving twin-network encoding (ITNE).

In ITNE, besides the connections between the input and output layers, interleaving connections are added for all hidden layer neurons between the two network copies. Specifically, for the $j$-th neuron of layer $i$, two variables, $\Delta y^{(i)}_j$ and $\Delta x^{(i)}_j$, are added to encode the distance of $y^{(i)}_j$ and $x^{(i)}_j$ between the two copies, where
\[
\Delta y^{(i)}_j = \hat{y}^{(i)}_j - y^{(i)}_j,
\]
\[
\Delta x^{(i)}_j = \hat{x}^{(i)}_j - x^{(i)}_j = relu(y^{(i)}_j + \Delta y^{(i)}_j) - relu(y^{(i)}_j).
\]

These additional distance variables reflect how different the two network copies are during the forward propagation. And the non-linear mapping from $\hat{y}^{(i)}_j$ to $\hat{x}^{(i)}_j$ is substituted by the mapping between $\Delta y^{(i)}_j$ and $\Delta x^{(i)}_j$. These changes enable the usage of the over-approximation techniques introduced below.

\subsection{Over-Approximation Techniques}
Leveraging ITNE, we design two over-approximation techniques, network decomposition (ND) and LP relaxation (LPR), to improve the global robustness certification efficiency. The two techniques are inspired by the local robustness certification work in~\cite{Huang_2020_EMSOFT}. However, there are major differences between global and local robustness certification, given that the former considers the entire input space $X$ while the latter only considers a given input $x^{(0)}\in X$, and our ND and LPR techniques are specifically designed for global robustness.

\paragraph{ITNE-based network decomposition (ND)}

The main idea of ND is to divide a neural network into sub-networks and decompose the entire optimization problem into smaller problems. The input of a sub-network is either the network input or the output of other sub-networks. Given the range of the input of a sub-network, the range of its output can be derived by solving an optimization problem of the sub-network, which can then be used as the input range of the next sub-network. As the complexity of MILP is exponential to the neural network size, using ND can significantly reduce the complexity of the optimization problem.
For global robustness certification, we also consider two copies of each sub-network. However, instead of finding the output range of each sub-network copy, we look for the range of one sub-network copy and the range of the output distance, based on the ITNE.

More specifically, a $w$-layer sub-network with input $x^{(i-w)}$ and output $x^{(i)}_j$ ($i \geq w$) is denoted as $F_w(x^{(i)}_j)$. For any variable $v$, its range is denoted as $\underline{\overline{v}}=[\underline{v}, \overline{v}]$.
Under ITNE, for each sub-network $F_w(x^{(i)}_j)$, given the input range $\underline{\overline{x}}^{(i-w)}$ and input distance range $\Delta \underline{\overline{x}}^{(i-w)}$, we can derive the output range $\underline{\overline{x}}^{(i)}_j$ and output distance range $\Delta \underline{\overline{x}}^{(i)}_j$ by solving the optimization problem in Eq.~\eqref{equ:probformul} for this sub-network. 

\paragraph{ITNE-based LP relaxation (LPR)}

The idea of LPR is to relax the ReLU relation $x = \max(0, y)$ into linear constraints given the range of $y$ as $[\underline{y}, \overline{y}]$. When $\overline{y} \leq 0$ or $\underline{y}\geq 0$, the ReLU relation degenerates to $x = 0$ or $x=y$. Otherwise, ReLU relation can be relaxed by three linear inequations:
\begin{equation}\label{equ:lprelax}
    x\geq 0,\quad
x \geq y,\quad
x \leq \frac{\overline{y}(y-\underline{y})}{\overline{y}-\underline{y}} 
\end{equation}

The ReLU relations of a neuron in the two network copies are formulated in Eqs.~\eqref{equ:milp_x} and~\eqref{equ:milp_x_2}.
In our work, instead of relaxing the ReLU relations in two network copies separately, we relax Eq.~\eqref{equ:milp_x} by the original LPR in Eq.~\eqref{equ:lprelax}, and relax the ReLU distance relation based on ITNE:
\begin{equation}\label{equ:reluDistance}
    \Delta x = relu(y + \Delta y) - relu(y).
\end{equation}

\begin{figure}[t]
\centering
\includegraphics[width=0.99\columnwidth]{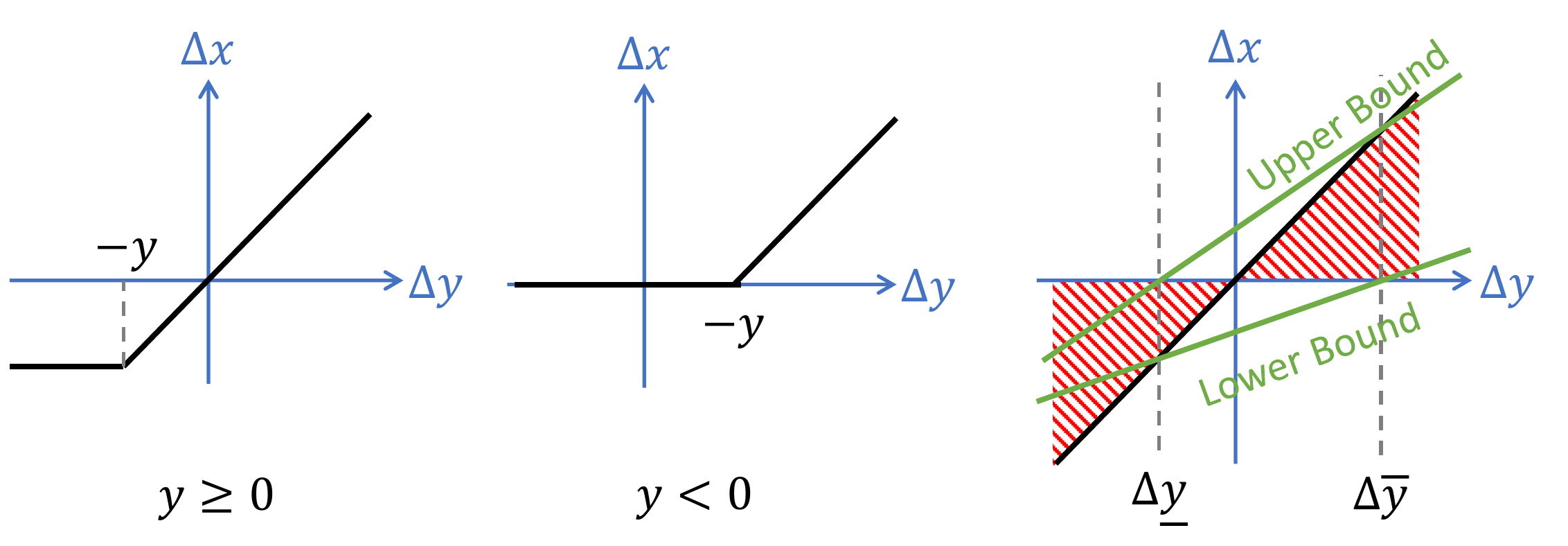}
\caption{Left: ReLU distance relation when $y\geq 0$; Middle: ReLU distance relation when $y < 0$; Right: LP-relaxation of ReLU distance relation. The ReLU distance relation for $\forall y\in\mathbb{R}$ lays in the shadowed area. Within the distance range $\Delta y\in[\Delta \underline{y}, \Delta \overline{y}]$, $\Delta x$ is bounded by the lower and upper bounds. }
\label{fig:lp_relax_dist}
\vspace{-12pt}
\end{figure}

For a given $y$, the relation between $\Delta x$ and $\Delta y$ is shown in the first two plots of Fig.~\ref{fig:lp_relax_dist}. Thus, the shadowed area in the third plot of Fig.~\ref{fig:lp_relax_dist} can cover all possible $(\Delta x, \Delta y)$ mappings for $\forall y\in \mathbb{R}$. Given $\Delta y\in[\Delta \underline{y}, \Delta \overline{y}]$, the relation between $\Delta x$ and $\Delta y$ can be bounded by the linear lower and upper bounds as shown in the third plot of Fig.~\ref{fig:lp_relax_dist}. Formally, the bound is
\begin{equation} \label{equ:lprelax_dist}
    \frac{l(u - \Delta y )}{u - l} \leq \Delta x \leq \frac{u(\Delta y - l)}{u - l}, 
\end{equation}
where $l =\min(0, \Delta \underline{y}$) and $u = \max(0, \Delta \overline{y})$.

\subsection{Illustrating example}
 
Consider the example neural network in Fig.~\ref{fig:NNstruct}. Assume that the input perturbation bound as $\delta = 0.1$  and the input domain as $x^{(0)}\in [-1, 1]^2$. The example neural network can be decomposed into three sub-networks: 
\[x^{(1)}_1 = relu(y^{(1)}_1) = relu(x^{(0)}_1 + 0.5x^{(0)}_2)\]
\[x^{(1)}_2 = relu(y^{(1)}_2) = relu(-0.5x^{(0)}_1 + x^{(0)}_2)\]
\[x^{(2)} = relu(y^{(2)}) = relu(x^{(1)}_1 - x^{(1)}_2).\] 

\begin{figure}[t]
    \centering
    \includegraphics[width=0.9\columnwidth]{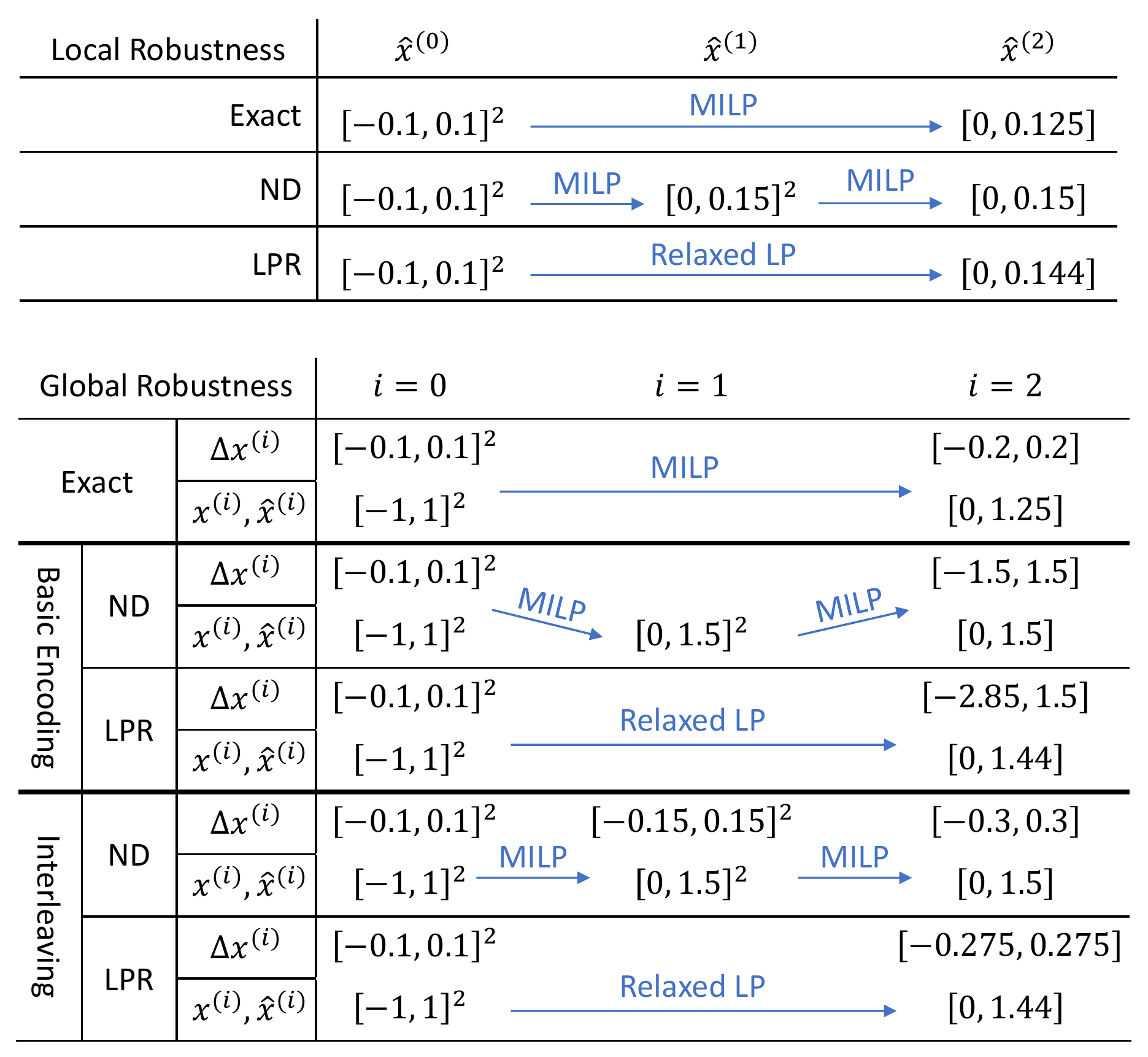}
    \caption{Illustrating example: The local and global robustness certification processes of exact MILP, network decomposition (ND) and LP relaxation (LPR). For global robustness, the ND and LPR for both basic and interleaving twin-network encoding (i.e., BTNE and ITNE) are illustrated.}
    \label{fig:motivateExp}
\end{figure}

For local robustness, with input $x^{(0)} = [0, 0]$, the certification processes for different techniques are shown in Fig.~\ref{fig:motivateExp}. 
The exact bound of $\hat{x}^{(2)}$ is derived by solving the MILP problem for the entire network. ND solves the MILP problems of the first two sub-networks to derive the bound of $\hat{x}^{(1)}$ and then derive the bound of $\hat{x}^{(2)}$ by solving the MILP problem of the third sub-network.
For LPR, given $\hat{y}^{(1)}_1, \hat{y}^{(1)}_2, \hat{y}^{(2)}\in[-0.15, 0.15]$, all ReLU relations can be relaxed based on Eq.~\eqref{equ:lprelax}, and the bound of $\hat{x}^{(2)}$ can be derived by solving the relaxed LP problem. As we can see, ND and LPR can provide tight (1.2x and 1.15x) over-approximations for local robustness.

For global robustness, under BTNE, there are no distance variables for hidden neurons. Thus, ND and LPR can only be applied to each individual network copy. ND will only find the range of $x^{(1)}$ and $\hat{x}^{(1)}$ and for the third sub-network, the distance information between two network copies is lost, resulting in a 7.5x over-approximation of the range of $\Delta x^{(2)}$. The LPR is based on the bounds $y^{(1)}, \hat{y}^{(1)} \in [-1.5, 1.5]^2$, and $y^{(2)}, \hat{y}^{(2)}\in[-1.5, 1.5]$, resulting in a 10.9x over-approximation.
On the other hand, under ITNE, the ITNE-based ND can first find the range of $x^{(1)}$ and $\Delta x^{(1)}$ and then derive the range of $\Delta x^{(2)}$, which is only 1.5x of the exact one. Given $\Delta y^{(1)}\in [-0.15, 0.15]^2$ and $\Delta y^{(2)}\in [-0.3, 0.3]$, the ITNE-based LPR derives a tight 1.38x over-approximation of the range of $\Delta x^{(2)}$. This shows that combing ITNE with ND and LPR significantly improves the approximation tightness over BTNE.

\subsection{Efficient Global Robustness Over-Approximation Algorithm} \label{sec:milp}

Finally, we design our global robustness certification algorithm by leveraging the ITNE encoding as well as the ND and LPR techniques, as shown in Algorithm~\ref{alg:algorithm}. Note that the MILP problems for sub-networks from ND are relaxed by LPR.
The pre-assigned window size $W$ is defined as the desired depth of sub-networks. 
For LPR, besides the range of the input and distance, $\underline{\overline{y}}^{(i-k)}$ and $\Delta \underline{\overline{y}}^{(i-k)}$ for $\forall k\in[0,w-1]$ are also needed. To evaluate $\underline{\overline{y}}^{(i)}_j$ and $\Delta \underline{\overline{y}}^{(i)}_j$, we can consider a $w$-layer sub-network with $y^{(i)}_j$ as the output (i.e., no ReLU activation at the output layer), denoted as $F_w(y^{(i)}_j)$. 
Formally, we have two types of sub-network optimization problems: $LpRelaxY$ and $LpRelaxX$, which is respectively encoded (lines 7 and 9) as ITNE of sub-network $F_w(y^{(i)}_j)$ and $F_w(x^{(i)}_j)$ (decomposed from network $F$ in lines 6 and 9). 
Both problems require the sub-network input range $\underline{\overline{x}}^{(i-w)}$, $\Delta \underline{\overline{x}}^{(i-w)}$ and the ranges for hidden neurons, i.e., $\underline{\overline{y}}^{(i-k)}$ and $\Delta \underline{\overline{y}}^{(i-k)}, \forall k\in[1,w-1]$ as prerequisites, while $LpRelaxX$ (line 11) also needs $\underline{\overline{y}}^{(i)}_j, \Delta \underline{\overline{y}}^{(i)}_j$, which is derived from $LpRelaxY$ (line 8).
Therefore, the ranges need to be evaluated layer by layer, by treating these hidden layer neurons as outputs of sub-networks $F_w(y^{(i)}_j)$ and $F_w(x^{(i)}_j)$. In such a way, when evaluating the ranges of certain layer, all ranges of previous layers have been derived.

\begin{algorithm}[tb]
\caption{Global Robustness Certification Algorithm}
\label{alg:algorithm}
\textbf{Input}: Neural Network $F$, window size $W$, refine param $r$\\
\textbf{Input}: input domain $X$, input perturbation bound $\delta$\\
\textbf{Output}: output variation bound $\bar{\varepsilon}$
\begin{algorithmic}[1] 
\STATE $\underline{\overline{x}}^{(0)} = X$
\STATE $\Delta \underline{\overline{x}}^{(i)} = [-\delta, \delta]^{m_0}$
\FOR{$i=1:n$}
\STATE $w = \max(i, W)$
\FOR{$j=1:m_i$}
\STATE $F_w(y^{(i)}_j)\leftarrow NetDecompose(F, y^{(i)}_j, w)$
\STATE $e_y = InterleaveTwinNetEncode(F_w(y^{(i)}_j))$
\STATE  ($\underline{\overline{y}}^{(i)}_j, \Delta \underline{\overline{y}}^{(i)}_j) \leftarrow LpRelaxY(e_y, r)$
\STATE $F_w(x^{(i)}_j)\leftarrow NetDecompose(F, x^{(i)}_j, w)$
\STATE $e_x = InterleaveTwinNetEncode(F_w(x^{(i)}_j))$
\STATE ($\underline{\overline{x}}^{(i)}_j, \Delta \underline{\overline{x}}^{(i)}_j) \leftarrow LpRelaxX(e_x, r)$
\ENDFOR
\ENDFOR
\STATE $\bar{\varepsilon} \leftarrow \max(|\Delta \underline{x}^{(n)}|, |\Delta \overline{x}^{(n)}|)$
\end{algorithmic}
\end{algorithm}

\textit{Selective Refinement:}
While LPR can remove all integer variables in the MILP formulation to reduce the complexity, such extreme over-approximation may be too inaccurate. Thus, we try to selectively refine a limited number of neurons, by \emph{not} relaxing their ReLU relations. This is similar to the layer-level refinement idea in~\cite{Huang_2020_EMSOFT}, but with a focus on global robustness. Specifically, a score of LPR is evaluated for each neuron as the worst-case inaccuracy, and the top $r$ neurons will be refined. The score is defined as the maximum distance between the lower and upper bound of LPR, i.e., the score is $-\overline{y}\underline{y}/(\overline{y}-\underline{y})$ for Eq.~\eqref{equ:lprelax}, and $\max(|\Delta \underline{y}|,|\Delta \overline{y}|)$ for Eq.~\eqref{equ:lprelax_dist}. The number of neurons for refinement $r$ is a parameter that can be set in LPR. 

\textit{Complexity Analysis:} MILP's complexity is polynomial to the number of variables and exponential to the number of integer variables. In our approach, for each MILP, the number of variables is linear to the number of neurons in the sub-network, and the number of integer variables is linear to the number of neurons selected for refinement. The number of MILPs to solve is linear to the total number of neurons.

\section{Experimental Results}

We first evaluate our algorithm on various DNNs and compare its results with exact global robustness (when available) and an under-approximated global robustness. 
Then, we demonstrate the application of our approach in a case study of safety verification for a vision-based robotic control system, and show the importance of efficient global robustness certification for safety-critical systems that involve neural networks.

We implement our efficient global robustness certification algorithm in Python. All MILP/LP problems in the algorithm are solved by Gurobi~\cite{gurobi}. All the experiments are conducted on a platform with a 4-core 3.6 GHz Intel Xeon CPU and 16 GB memory. All neural networks are modeled in Tensorflow.

\subsection{Comparison with Other Methods}

\begin{table}[t]
\centering
\caption{Neural network setting and experimental results. }
\begin{tabular}{|c|c|c|c|c|c|c|c|}
\hline
Dataset & ID & Neurons & $t_R$ & $t_M$ & $t_{our}$ & $\varepsilon$ & $\overline{\varepsilon}$ (ours) \\ \hline
\multirow{4}{*}{\begin{tabular}[c]{@{}c@{}}Auto\\ MPG\end{tabular}} & 1 & 8 & 2s & 0.1s & 0.3s & 0.0583 & 0.0657 \\ \cline{2-8} 
 & 2 & 12 & 130s & 0.2s & 0.4s & 0.0527 & 0.0722 \\ \cline{2-8} 
 & 3 & 16 & 8h & 0.8s & 1s & 0.0496 & 0.0653 \\ \cline{2-8} 
 & 4 & 32 & $>24$h & 74s & 5s & 0.0481 & 0.0673 \\ \hline \hline
Dataset & ID & Neurons & \multicolumn{2}{c|}{Layers} & $t_{our}$ & $\underline{\varepsilon}$ & $\overline{\varepsilon}$ (ours) \\ \hline
\multicolumn{1}{|l|}{\begin{tabular}[c]{@{}l@{}}Auto\\ MPG\end{tabular}} & 5 & 64 & \multicolumn{2}{c|}{FC:3} & 50s & 0.0452 & 0.0731 \\ \hline
\multirow{6}{*}{MNIST} & \multirow{2}{*}{6} & \multirow{2}{*}{1416} & \multicolumn{2}{c|}{\multirow{2}{*}{\begin{tabular}[c]{@{}c@{}}Conv:1\\ FC:2\end{tabular}}} & \multirow{2}{*}{4.8h} & 0.347 & 0.578 \\ \cline{7-8} 
 &  &  & \multicolumn{2}{c|}{} &  & 0.300 & 0.572 \\ \cline{2-8} 
 & \multirow{2}{*}{7} & \multirow{2}{*}{3872} & \multicolumn{2}{c|}{\multirow{2}{*}{\begin{tabular}[c]{@{}c@{}}Conv:2\\ FC:2\end{tabular}}} & \multirow{2}{*}{3.3h} & 0.453 & 0.874 \\ \cline{7-8} 
 &  &  & \multicolumn{2}{c|}{} &  & 0.420 & 0.723 \\ \cline{2-8} 
 & \multirow{2}{*}{8} & \multirow{2}{*}{5824} & \multicolumn{2}{c|}{\multirow{2}{*}{\begin{tabular}[c]{@{}c@{}}Conv:3\\ FC:2\end{tabular}}} & \multirow{2}{*}{3.5h} & 0.519 & 1.521 \\ \cline{7-8} 
 &  &  & \multicolumn{2}{c|}{} &  & 0.407 & 1.175 \\ \hline
\end{tabular}
\label{tab:results}

{\raggedright In Table~\ref{tab:results}, `Layers' are the type and number of layers; `Neurons' is the total number of hidden neurons; 
$t_{R}$, $t_M$, $t_{our}$ are the certification time of exact MILP, Reluplex, and our approach, respectively; 
$\varepsilon$, $\underline{\varepsilon}$, $\overline{\varepsilon}$ are the exact (derived by Reluplex or MILP), under-approximated (derived by projected gradient descent (PGD) among dataset), and over-approximated (derived by our approach) output variation bounds, respectively.
\par}
\vspace{-12pt}
\end{table}

We conduct experiments on a set of DNNs with different sizes. 
The DNNs are trained on two datasets: the Auto MPG dataset~\cite{Auto_MPG} for vehicle fuel consumption prediction, and the MNIST handwritten digits classification dataset~\cite{MNIST}. Besides the fully-connected (FC) output layer, the Auto MPG DNNs contain 2 FC hidden layers with the number of hidden neurons ranging from 8 to 64. MNIST DNNs contain 1 to 3 convolutional layers followed by 1 FC hidden layer, with 1472 to 7176 hidden neurons. We certify the output variation bound $\varepsilon$ based on the input perturbation bound $\delta$, where $\delta=0.001$ for Auto MPG DNNs and $\delta=2/255$ for MNIST DNNs.

The network settings and the comparison among different approaches are shown in Table~\ref{tab:results}.
For our approach, window size $W=2$ for Auto MPG DNNs, and $W=3$ for MNIST DNNs. Half neurons are refined in Auto MPG DNNs and 30 neurons in each layer are refined in MNIST DNNs. We present 2 outputs of MNIST (out of 10) in Table~\ref{tab:results}, and others show similar differences between different methods. 

For small networks (DNN-1 to DNN-5), we compare our over-approximated output variation bound $\overline{\varepsilon}$ with the exact bound $\varepsilon$ solved by Reluplex~\cite{Katz_2017_CAV_Reluplex} (latest version that is  integrated in the tool Marabou~\cite{Katz_2019_CAV} is used) and the MILP encoding in Eq.~\eqref{equ:probformul}. We can see that the runtime of Reluplex $t_R$ and MILP $t_M$ quickly increases with respect to neural network size. None of them can get a solution within 24 hours for DNN-5 (with only 64 hidden neurons).
From DNN-1 to DNN-4, our algorithm can derive an over-approximated global robustness with much slower runtime increase, while only have about 13\% to 40\% over-approximation over the exact global robustness.

Starting from DNN-5, the two exact certification methods cannot find a solution in reasonable time. There is also no other work in the literature that can derive a sound and deterministic global robustness. To assess how good our over-approximated results are for larger networks, we evaluate an under-approximation of the global robustness, inspired by~\cite{Ruan_2019_IJCAI}. Specifically, for each data sample in the dataset, we leverage projected gradient descent (PGD)~\cite{Madry_2018_ICLR} to look for an adversarial example in the input perturbation bound that maximizes the output variation, and treat the maximal output variation among the entire dataset as an \emph{under-approximated} output variation bound $\underline{\varepsilon}$.  
The exact global robustness should be between the under-approximation $\underline{\varepsilon}$ derived from this dataset-wise PGD, and the over-approximation $\overline{\varepsilon}$ derived by our certification algorithm.
The experiments from DNN-6 to DDN-8 demonstrate that \textbf{our method can provide meaningful over-approximation (less than 3x of the under-approximation) for DNNs with more than 5000 hidden neurons within 5 hours}.

\subsection{Case Study on Control Safety Verification}

For control systems that use neural networks for perception, a critical and yet challenging question is whether the system can remain safe when there is perturbation/disturbance to the perception neural network input.  
In~\cite{Wang_2021_DATE}, the authors formulate this as a design-time safety assurance problem based on global robustness (although did not provide a solution), i.e., if a global robustness bound $\epsilon$ can be derived for an input perturbation bound $\delta$, such $\epsilon$ can be viewed as the state estimation error bound in control and leveraged for control safety verification.

In this section, leveraging our approach for bounding global robustness, we conduct a case study for safety verification of control systems that use neural networks for perception. We consider an advanced cruise control (ACC) scenario where an ego vehicle is following a reference vehicle. The ego vehicle is equipped with an RGB camera, and a DNN is designed to estimate the distance from the reference vehicle based on the camera image. We assume that the captured image may be slightly perturbed. A feedback controller takes the estimated distance, along with the ego vehicle speed, to generate the control input, which is the acceleration of the ego vehicle. 

\begin{figure}[t]
\centering
\includegraphics[width=0.8\columnwidth]{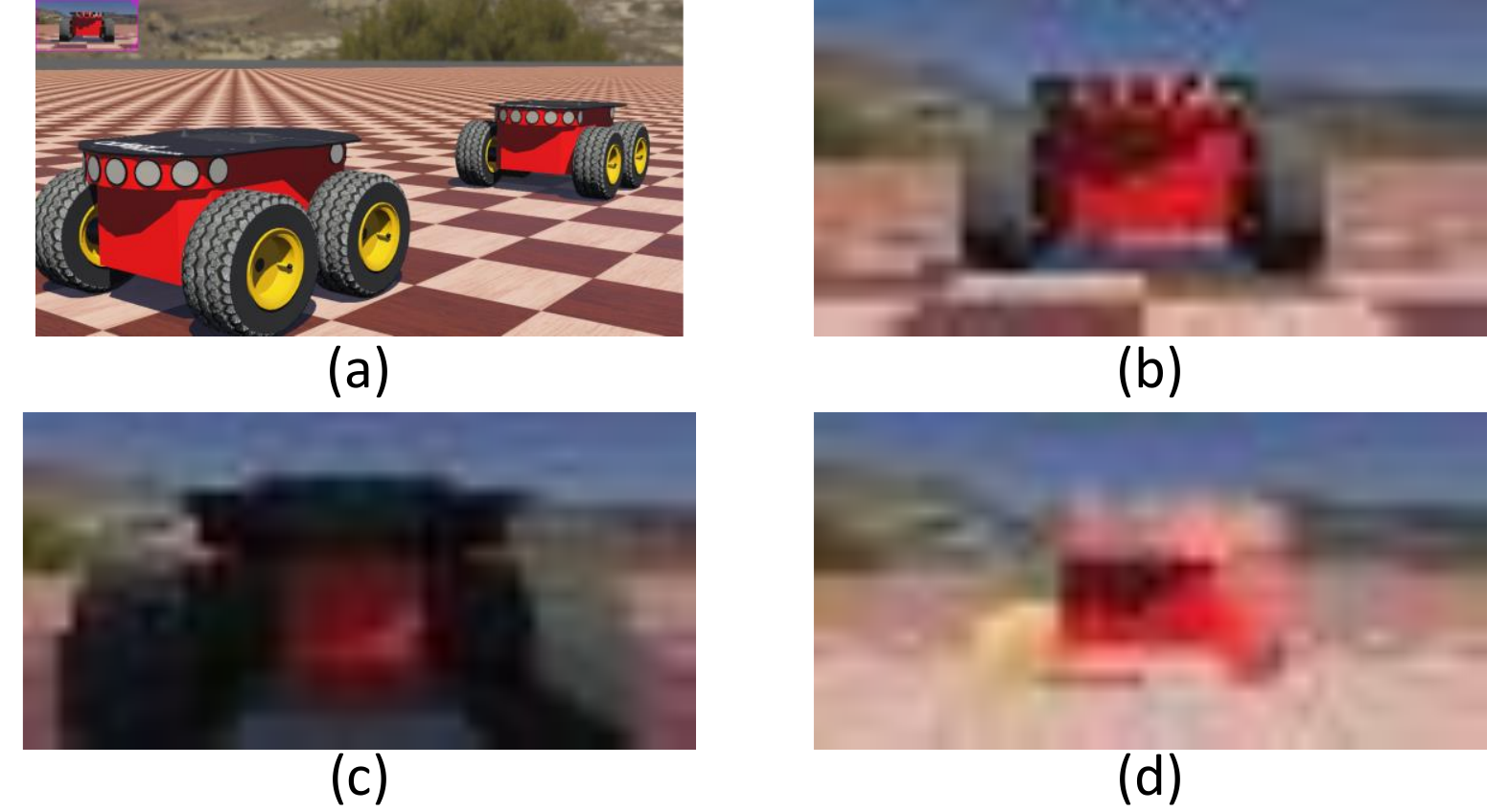}
\caption{(a) Webots simulation for case study, where ego vehicle (left) follows reference vehicle (right); (b) An example image captured by ego vehicle; (c) Lower bound of DNN input space; (d) Upper bound of DNN input space.}
\label{fig:webots}
\vspace{-12pt}
\end{figure}

We model this example in the tool Webots~\cite{webots} (Fig.~\ref{fig:webots}). The ego vehicle is safe if distance $d\in[0.5, 1.9]$ and speed $v_e\in[0.1,0.7]$. The reference vehicle speed $v_r$ is randomly adjusted within $[0.2, 0.6]$. 
The sampling period is $100$ms.
The camera takes RGB images with resolution $24\times48$. 
A 5-layer (3 convolutional and 2 FC) DNN  is trained with 100k pre-captured images. The system dynamics is modeled as:
\[
x[k+1] = 
\begin{bmatrix}
1 & -0.1\\
0 & 1
\end{bmatrix}
x[k] + 
\begin{bmatrix}
-0.005\\
0.1
\end{bmatrix}
u[k] + 
\begin{bmatrix}
1\\
0
\end{bmatrix}
w_1[k] + w_2[k]
\]
where the system state $x = [d - 1.2, v_e - 0.4]^\top$ contains the normalized distance and speed. $w_1 = 0.4 - v_r$ is the external disturbance, where $v_r$ is bounded within $[0.2, 0.6]$ as previously mentioned. $w_2 = [w_d, w_v]^\top$ is the inaccuracy of this system dynamics model. The control input follows the feedback control law $u=K\hat{x}$, where $K = [0.3617,-0.8582]$. $\hat{x}$ is the estimated system state, and the state estimation error is $\Delta x = \hat{x} - x$. The estimation error of ego vehicle speed $v_e$ is assumed as $0$. The bound of $w_d$, $w_v$ are profiled based on simulations, and bounded as $|w_d|\leq 5e-4$ and $|w_v|\leq 3e-5$. The estimation error for distance contains two parts $\Delta d = \Delta d_1 + \Delta d_2$, where $\Delta d_1$ is the error coming from the DNN model inaccuracy, while $\Delta d_2$ is the output variation caused by input perturbation.

The DNN model inaccuracy $\Delta d_1$ is determined by the worst-case model inaccuracy among the dataset and set as $|\Delta d_1|\leq 0.0730$. The DNN output variation $\Delta d_2$ is the focus of this study and can be bounded by our global robustness certification algorithm. 
Specifically, the input perturbation in this study is bounded by $\delta = 2/255$ (Fig.~\ref{fig:webots} illustrates the upper and lower bound of each pixel for the input space). Then, using our approach, we can derive a certified output variation bound $|\Delta d_2|\leq \overline{\epsilon} = 0.0568$. Combined with $\Delta d_1$, we have $|\Delta d|\leq 0.0730 + 0.0568 = 0.1298$. 

With bounds on $\Delta d$ and other variables, the vehicle control safety can be verified based on the computation of control invariant set, similarly as in~\cite{huang2020opportunistic}. 
Through such invariant set based verification, we find that as long as the distance estimation error $\Delta d$ is within $[-0.14, 0.14]$, the system will always be safe. Since our global robustness certification bounds $\Delta d \leq 0.1298$, we can assert that our ACC system is safe with this DNN design under the assumed perturbation bound.

We also deploy our DNN model in Webots and add adversarial perturbation on the input images by Fast Gradient Sign Method (FGSM)~\cite{Goodfellow_2015_ICLR}. With more than 1000 minutes of simulation, we find that when the perturbation bound is $\delta = 2/255$ as assumed, the distance estimation error $\Delta d$ never exceeds the $[-0.14, 0.14]$ bound and the system is always safe (as expected since it was formally proven). When the perturbation bound $\delta$ is increased to $5/255$, $\Delta d$ sometimes exceeds the bound, although unsafe system state is not observed. When $\delta$ is increased to $10/255$, about 17\% of the simulations encounter unsafe states. This shows the impact of input perturbation on system safety and the importance of our global robustness analysis.

\section{Conclusion}

We present an efficient certification algorithm to provide a sound and deterministic global robustness for neural networks with ReLU activation.
Experiments demonstrate that our approach is much more efficient and scalable than exact global robustness certification approaches while providing tight over-approximation, and a case study further demonstrates the application of our approach in safety-critical systems. Our future work will continue improving the approach's efficiency, possibly via parallelization on multi-core platforms, to enable its usage for larger-size perception neural networks.

\bibliographystyle{IEEEtran}
\bibliography{IEEEabrv,date22.bib}

\end{document}